%% file: example_paper.tex
\documentclass{article}

\usepackage{microtype}
\usepackage{graphicx}
\usepackage{subcaption}
\usepackage{booktabs} % for professional tables
\usepackage{balance}
\usepackage{hyperref}

\usepackage{siunitx}

\usepackage{comment}

\usepackage{xspace}
\newcommand{\aname}{SeerAttention\xspace}
\newcommand{\gate}{AttnGate\xspace}
\newcommand{\gates}{AttnGates\xspace}
\newcommand{\Figref}[1]{\cref{#1}}
\newcommand{\tabref}[1]{Table~{\ref{#1}}}

\usepackage{multirow}
\usepackage{graphicx}
\usepackage{xcolor}

\usepackage{url}
\usepackage[linesnumbered,ruled,vlined]{algorithm2e}

\usepackage[accepted]{icml2025}

\usepackage{amsmath}
\usepackage{amssymb}
\usepackage{mathtools}
\usepackage{amsthm}

\usepackage[capitalize,noabbrev]{cleveref}

\theoremstyle{plain}

\theoremstyle{definition}

\theoremstyle{remark}

\usepackage[textsize=tiny]{todonotes}

\icmltitlerunning{SeerAttention: Learning Intrinsic Sparse Attention in Your LLMs}

\begin{document}

\twocolumn[
\icmltitle{SeerAttention: Learning Intrinsic Sparse Attention in Your LLMs}

% It is OKAY to include author information, even for blind
% submissions: the style file will automatically remove it for you
% unless you've provided the [accepted] option to the icml2025
% package.

% List of affiliations: The first argument should be a (short)
% identifier you will use later to specify author affiliations
% Academic affiliations should list Department, University, City, Region, Country
% Industry affiliations should list Company, City, Region, Country

% You can specify symbols, otherwise they are numbered in order.
% Ideally, you should not use this facility. Affiliations will be numbered
% in order of appearance and this is the preferred way.

% \author{%
%   \textbf{Yizhao Gao}\textsuperscript{1}\thanks{Equal Contribution}\hspace{4pt}\thanks{Work done during the internship at Microsoft Research} 
%   \quad
%   \textbf{Zhichen Zeng}\textsuperscript{2}\footnotemark[1]\hspace{4pt}\footnotemark[2] 
%   \quad
%   \textbf{Dayou Du}\textsuperscript{3}\footnotemark[2] 
%   \quad
%   \textbf{Shijie Cao}\textsuperscript{4}\thanks{Corresponding author: shijiecao@microsoft.com}\\
%   \textbf{Hayden Kwok-Hay So\textsuperscript{1} 
%   \quad
%   \textbf{Ting Cao}\textsuperscript{4} 
%   \quad
%   \textbf{Fan Yang}\textsuperscript{4}
%   \quad
%   \textbf{Mao Yang}\textsuperscript{4}}\\
%   \textsuperscript{1}University of Hong Kong 
%   \quad
%   \textsuperscript{2}University of Washington \\
%   \textsuperscript{3}Hong Kong University of Science and Technology (Guangzhou)\\
%   \textsuperscript{4}Microsoft Research \\
% }

\icmlsetsymbol{equal}{*}

\begin{icmlauthorlist}
\icmlauthor{Yizhao Gao}{equal,hku}
\icmlauthor{Zhichen Zeng}{equal,uw}
\icmlauthor{Dayou Du}{hkust}
\icmlauthor{Shijie Cao}{ms}
\icmlauthor{Peiyuan Zhou}{nv}
\icmlauthor{Jiaxing Qi}{nv}
\icmlauthor{Junjie Lai}{nv}
%\icmlauthor{}{sch}
\icmlauthor{Hayden Kwok-Hay So}{hku}
\icmlauthor{Ting Cao}{ms}
\icmlauthor{Fan Yang}{ms}
\icmlauthor{Mao Yang}{ms}
%\icmlauthor{}{sch}
%\icmlauthor{}{sch}
\end{icmlauthorlist}

\icmlaffiliation{hku}{The University of Hong Kong}
\icmlaffiliation{uw}{University of Washington}
\icmlaffiliation{hkust}{Hong Kong University of Science and Technology (Guangzhou)}
\icmlaffiliation{ms}{Microsoft Research}
\icmlaffiliation{nv}{NVIDIA}

% \icmlaffiliation{comp}{Company Name, Location, Country}
% \icmlaffiliation{sch}{School of ZZZ, Institute of WWW, Location, Country}

\icmlcorrespondingauthor{Shijie Cao}{shijiecao@microsoft.com}
% \icmlcorrespondingauthor{Firstname2 Lastname2}{first2.last2@www.uk}

% You may provide any keywords that you
% find helpful for describing your paper; these are used to populate
% the "keywords" metadata in the PDF but will not be shown in the document
\icmlkeywords{Machine Learning, ICML}

\vskip 0.3in
]

% this must go after the closing bracket ] following \twocolumn[ ...

% This command actually creates the footnote in the first column
% listing the affiliations and the copyright notice.
% The command takes one argument, which is text to display at the start of the footnote.
% The \icmlEqualContribution command is standard text for equal contribution.
% Remove it (just {}) if you do not need this facility.

% \printAffiliationsAndNotice{}  % leave blank if no need to mention equal contribution
\printAffiliationsAndNotice{\icmlEqualContribution} % otherwise use the standard text.

\begin{abstract}
Attention is the cornerstone of modern Large Language Models (LLMs). Yet its quadratic complexity hinders efficiency and scalability, especially for long-context processing. A promising approach is to leverage sparsity in attention. 
However, existing sparsity-based solutions predominantly rely on \textit{predefined patterns or heuristics} at the attention head level, struggling to adapt dynamically to different contexts efficiently.

We propose \aname{}, a simple yet effective attention mechanism that directly learns the block-level attention sparsity from the LLM itself.
Inspired by the gating mechanism in Mixture of Experts (MoE), \aname{} augments the conventional attention with a \textbf{learnable gate} that \textbf{selectively activates} \textbf{important blocks} within the attention map.
Specifically, the gate first pools the query (Q) and key (K) tensors along the sequence dimension and processes them through learnable linear layers. The resulting matrices are then multiplied together to produce the gating scores, which are used to predict block-level attention sparsity.
Combined with our block-sparse FlashAttention kernel, \aname{} can achieve significant speedup on GPUs. 
When applied to pre-trained LLMs, \aname{} only requires training the gate parameters in a lightweight self-distillation manner, allowing rapid convergence.
Our evaluation results demonstrate that \aname{} achieves better model accuracy and lower latency for long-context pre-filling compared to prior methods. Code is available at: \url{https://github.com/microsoft/SeerAttention}
%compared full attention, speedup.

%with xxx calibration data (~ 64k, 500 step, 16 batch size, 0.5B tokens, 40 A100 hours), which helps the gate learn to effectively identify and prioritize important contextual relationships.

\end{abstract}

\input{01-intro}
\input{02-background}

\input{03-method}

\input{04-results}

\input{05-conclusion}
\balance
\bibliography{example_paper}
\bibliographystyle{icml2025}

%%%%%%%%%%%%%%%%%%%%%%%%%%%%%%%%%%%%%%%%%%%%%%%%%%%%%%%%%%%%%%%%%%%%%%%%%%%%%%%
%%%%%%%%%%%%%%%%%%%%%%%%%%%%%%%%%%%%%%%%%%%%%%%%%%%%%%%%%%%%%%%%%%%%%%%%%%%%%%%
% APPENDIX
%%%%%%%%%%%%%%%%%%%%%%%%%%%%%%%%%%%%%%%%%%%%%%%%%%%%%%%%%%%%%%%%%%%%%%%%%%%%%%%
%%%%%%%%%%%%%%%%%%%%%%%%%%%%%%%%%%%%%%%%%%%%%%%%%%%%%%%%%%%%%%%%%%%%%%%%%%%%%%%
\newpage
\appendix
\onecolumn
% \section{Appendix}

\input{00-appendix}

% You can have as much text here as you want. The main body must be at most $8$ pages long.
% For the final version, one more page can be added.
% If you want, you can use an appendix like this one.  

% The $\mathtt{\backslash onecolumn}$ command above can be kept in place if you prefer a one-column appendix, or can be removed if you prefer a two-column appendix.  Apart from this possible change, the style (font size, spacing, margins, page numbering, etc.) should be kept the same as the main body.
%%%%%%%%%%%%%%%%%%%%%%%%%%%%%%%%%%%%%%%%%%%%%%%%%%%%%%%%%%%%%%%%%%%%%%%%%%%%%%%
%%%%%%%%%%%%%%%%%%%%%%%%%%%%%%%%%%%%%%%%%%%%%%%%%%%%%%%%%%%%%%%%%%%%%%%%%%%%%%%

\end{document}

%% file: 01-intro.tex
\section{Introduction}

Attention is a fundamental mechanism in transformer-based LLMs~\citep{attention}. Despite its effectiveness, the quadratic complexity of attention demands substantial computation and memory resources, limiting the scalability and efficiency of LLMs, especially for long-context windows. This challenge has become an active research topic in the community. One potential solution is to replace the quadratic attention with cheaper architectures like linear attention or recurrent networks~\citep{linearattention, mamba, rwkv} with subquadratic complexity.
While these approaches are more efficient, the majority of state-of-the-art large language models (LLMs) continue to use full attention to achieve better performance.
% Although efficient, these solutions struggle to match the efficacy of full attention, especially when the scale is large.

A promising approach with increasing interests is to leverage sparsity in attention. Sparsity commonly exists in attention maps, and it becomes more prominent in longer contexts. In certain LLM attention heads, the sparsity ratio can reach 95\% or even 99\%, posing great opportunities for efficiency improvements. However, prior studies often rely on predefined sparsity patterns or heuristics to approximate the attention mechanism~\citep{minference, moa, hip, sample_attn, hyperattention, duo}.
The sparsity observed in attention maps varies significantly across different models, input contexts and attention heads, making predefined patterns or heuristics insufficient.

In this paper, we introduce \aname{}, a simple yet effective attention mechanism that directly learns the intrinsic attention sparsity from the LLM itself, without relying on predefined sparsity patterns.
To achieve this, \aname{} augments conventional attention with a learnable gate that selectively activates a small subset of important blocks in the attention map, drawing inspiration from the gating mechanism in MoE~\cite{moe}.
The gating process in \aname{} consists of three key steps. First, the query ($\mathbf{Q}$) and key ($\mathbf{K}$) matrices are pooled along the sequence length to reduce the substantial gating cost while preserving essential information. Second, the pooled $\mathbf{Q}$ and $\mathbf{K}$ representations are processed through linear layers to enable learnability. Finally, the transformed representations are multiplied to compute gating scores, which adaptively identify the most important blocks in the attention map.
By skipping unimportant blocks, the resulting block-sparse attention significantly reduces both I/O overhead and computational cost.
 
To train the gating mechanism in \aname{}, we adopt a lightweight self-distillation approach. Specifically, the pooled attention map from standard attention serves as the teacher, guiding block-level sparsity learning in the \aname{} gate, which acts as the student.
Importantly, for pre-trained LLMs, \aname{} only requires learning the gating parameters, while all other model parameters remain fixed. This leads to fast training process as only the newly added gate weights requires to compute gradient. For instance, when applied to a Llama-3.1-8B-Instruct model, \aname{} uses only 0.5B tokens for gate distillation, which is approximately 40 A100 GPU hours for training.

% add some real numbers
Our results demonstrate that \aname{} surpasses state-of-the-art sparse attention methods like Minference~\citep{minference}, MoA~\citep{moa} and DuoAttention~\citep{duo} in terms of long context model accuracy and pre-filling latency. 
\aname{} achieves highly linear speedup over dense configurations, delivering a 7.3× speedup with 90\% sparsity on sequences of 128k.
Notably, in contrast to previous methods that require careful calibration of sparse configuration for different heads, \aname{} offers strong capabilities of adaptation to different heads and contexts. 
Remarkably, on top of block-sparse pattern, \aname{} exhibits the ability to learn more diverse patterns, including A-shape and Vertical-Slash, further demonstrating its versatility and performance.

Our contributions can be summarized as follows:
\begin{itemize}
    \item We propose \aname, an innovative learnable attention gating mechanism to enhance efficiency for long-context LLMs.
    \item We have developed a self-distillation training scheme to efficiently train the \gate, enabling it to learn the intrinsic sparsity of a pre-trained model.
    \item Experiments show that \aname outperforms previous approaches, offering adaptability to various context lengths and sparsity ratios.
\end{itemize}

%% file: 02-background.tex
\section{Background and Related Works}

\paragraph{Powerful but Complex Attention in Transformer.}

%attention is fundamental to transfomr based llms. but with quadratic complexity especially in the era of long context llm. realted works: linear attention and rnn based architecture.

The advent of attention mechanisms, particularly within the Transformer architecture~\citep{attention}, marked a significant advancement in natural language processing. 
Attention enables improved handling of long-range dependencies and a better understanding of context by
attending each token to every other token in the sequence,  resulting in a quadratic memory and time complexity \(O(n^2)\), where \(n\) is the sequence length.
This presents a significant challenge as the community moves towards LLMs that can process increasingly longer contexts.
Many studies explore alternative attention mechanisms to mitigate this complexity.
The Reformer architecture~\citep{reformer} reduces the complexity to \(O(n \log n)\) and the linear attention mechanism~\citep{linearattention,yang2023gated} further decreases complexity to \(O(n)\).
%The Reformer architecture~\citep{reformer} reduces the complexity to \(O(n \log n)\) by using locality-sensitive hashing. The linear attention mechanism~\citep{linearattention} further decreases complexity to \(O(n)\) by using a linear dot product of kernel features and the associativity of matrix products.
Recently, there has been a trend of revisiting recurrent neural networks, leading to the proposal of new architectural frameworks such as RWKV~\citep{rwkv}, RetNet~\citep{retnet}, and Mamba~\citep{mamba}.
Despite their promise of efficiency, these methods struggle to match the performance of full attention mechanisms, particularly with larger models and longer contexts.
%\subsection{Long Context}
%Model based. Mamba
%System based. Flash-attention

\paragraph{Intrinsic but Dynamic Sparsity in Attention.}

Attention mechanisms inherently exhibit sparsity, which arises from the attention map $\mathbf{A}$ generated by $\mathbf{Q}$ and $\mathbf{K}$: $\mathbf{A} = \mathrm{softmax}(\mathbf{QK^{T}}/\sqrt{d})$.
%\begin{equation}
%    \mathbf{A} = \mathrm{softmax}(\mathbf{QK^{T}}/\sqrt{d})
%\end{equation}
The softmax function often produces a multitude of negligible scores that can be treated as zeros without impacting model accuracy~\citep{zaheer2020big, xieyuan_sparse, spatten, spasretrans, liu2023deja}.
Attention sparsity becomes more pronounced with longer contexts, presenting opportunities to optimize inference speed.
%Identifying the sparsity within attention maps could accelerate multiplications of queries and keys (as output sparsity) and the subsequent multiplications with values (as input sparsity) post-softmax.
Unfortunately, this sparsity is dynamic, varying across different context inputs and attention heads, each displaying distinct sparsity locations and ratios. 
Prior research has attempted to approximate attention sparsity using predefined patterns and heuristics~\citep{moa,minference} for different attention heads.
Yet, these methods lack generality and often rely on handcrafted features, struggling to fully capture the sparsity behavior of attention mechanisms.
The dynamic and input-dependent nature of attention sparsity echoes the principles of Mixture of Experts (MoE) models ~\citep{moe,MoEswitch} suggesting that sparsity should ideally be learned directly from data within the model itself. This approach would allow models to adaptively harness sparsity, improving efficiency while maintaining accuracy.

%Unfortunately, since $\mathbf{A}$ is output sparse, obtaining this fine-grained attention map itself already costs significant computation and I/O. Therefore, a more practical approach to leveraging attention sparsity is to enable the ability to also reduce the unnecessary computation and I/O related to $\mathbf{Q}$ and $\mathbf{K}$ beforehand. Previous works employing static or heuristic-based sparse attention have made important strides in addressing these issues, but they often lack generalizability and rely on handcrafted features. 

\paragraph{Long-Context LLM Optimizations.}
%1.long context is important
%2.long context is challenge: compute and memory (KV cache)
%3.many research directions to reduce the cost of long context.
%for example:
%1)optimize prefill: sparse attention, promot compression
%2)optimize decode: sparse load
%3)compress kv cache: kv cache sharing, kv eviction, kv quantization.

The ability to process long contexts is crucial for large language models (LLMs) as it enables them to retain and utilize more extensive information. However, it comes with substantial computational and memory costs. Various research efforts have explored different strategies to optimize long-context processing. One major direction is improving prefill efficiency, where techniques such as prompt compression~\citep{jiang2023longllmlingua,mu2024learning, chuang2024learning} and sparse attention~\citep{minference,moa,acharya2024star,zhang2024selective}.
Another approach focuses on optimizing the decoding phase by introducing sparse loading mechanisms~\citep{yang2024tidaldecode,chen2024magicpig}.
Additionally, several methods aim to compress the KV cache, including KV cache sharing~\citep{ainslie2023gqa,brandon2024reducing}, KV eviction policies~\citep{zhang2023h2o,li2024snapkv,ge2023model}, and KV quantization~\citep{liu2024kivi,hooper2024kvquant,dong2024qaq,zhang2024kv}.

%% file: 03-method.tex
\section{SeerAttention}
% we removed drawing insights from or motivated by MoE in abstract and introduction, but i think we can explicitly say this in design section.
%\subsection{Overall architecture} \label{sec: model arch}

\begin{figure*}[t]
    \centering
    \includegraphics[width=1\linewidth]{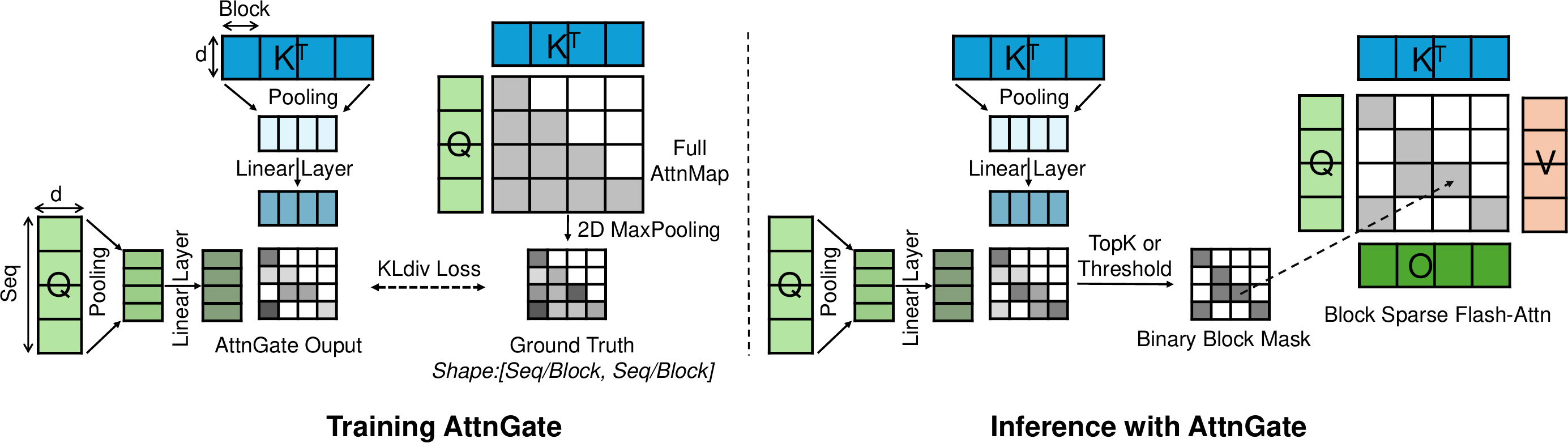}
    \caption{\textbf{Overall of SeerAttention.} The \gate in \aname first pools the Q and K tensors in sequence dimension and passes through learnable linear layers. The matrix multiplied results are then trained to mimic the 2D maxpooled results of the original pre-trained LLM in a self-distillation manner. During inference, the gating score can be used to predict the block-level attention sparsity through TopK or Thresholding. Noted that details gate design like pooling composition, RoPE and softmax are omitted in this diagram for simplicity.}
    \label{fig:SeerAttention}
\end{figure*}

\aname{} adopts a fully learning-based approach to adaptively identify attention sparsity in LLMs and leverages the learned sparsity for efficient inference. To ensure efficiency on modern hardware like GPUs, we focus on learning block sparsity, which can seamlessly integrate with the tiling computation scheme of FlashAttention~\citep{flash1, flash2}.
\Figref{fig:SeerAttention} illustrates the overall diagram of \aname, which augments conventional attention with a learnable gating module, termed \textit{Attention Gate} (\gate). The \gate modules contain learnable parameters (linear layers) and are distilled to mimic the 2D-Maxpooled results of the attention maps. 
At inference time, the \gate can predict the block-level sparsity for the subsequent attention computation with a block-sparse FlashAttention kernel, which significantly enhances performance by reducing I/O and computation overhead.

\subsection{Attention Gate Design}
The \gate module is designed to learn block-wise information with minimal overhead. It takes the original matrices $\mathbf{Q}$ and $\mathbf{K}$ as inputs and downsamples them using pooling operations along the sequence dimension. As shown in \Figref{fig:SeerAttention}, for a given attention head, the sizes of the pooled $\mathbf{Q}$ and $\mathbf{K}$ become $[seq/B, d]$, where $B$ is the kernel and stride size of the pooling operation (non-overlapped blocks). The downsampled $\mathbf{Q}$ and $\mathbf{K}$ are then processed through a linear layer and multiplied together, similar to the standard attention operation. This results in a matrix of size $[seq/B, seq/B]$, where each element corresponds to one block in the original full attention map. With a typical block size of 64, the output of the \gate module is only $\frac{1}{4096}$ the size of the original attention map, making it super efficient to compute. 
% During inference, by selecting the Top-k blocks in each row or thresholding the gating score, the block-sparse FlashAttention kernel can efficiently load and process only the active blocks. 
To its simplest form, the \gate computation can be expressed as:
\begin{equation}
score = \text{softmax} \left(
    \frac{(W_q \, P_q(Q)) \cdot (W_k \, P_k(K))^T}{\sqrt{d}}
\right)
\end{equation}
where $P_q$ and $P_k$ represents the pooling operations for $\mathbf{Q}$ and $\mathbf{K}$, and $d$ is the hidden size of the tensors similar to attention computation.

% Downsampling QK + Linear. 

% Training + softmax. 

% Inference no softmax + topk + sparse kernel.

% Different pooling method exploaration.

\paragraph{Pooling Method Selection.}
\begin{figure}[]
    \centering
    \includegraphics[width=1\linewidth]{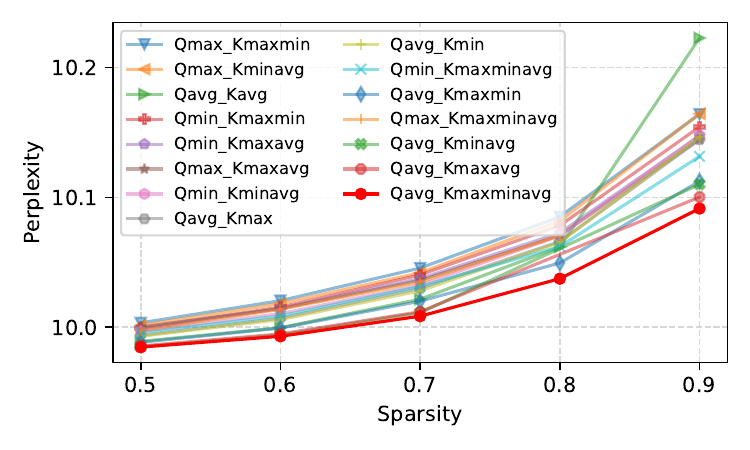}
    \caption{\textbf{Test Perplexing of Different Pooling Method Combinations on PG19.} The best configuration observed is using AvgPooling on Q and a combination of Max, Min, AvgPooling on K.}
    \label{fig:pooling}
\end{figure}
As pooling operations downsample the tensors and might lead to information loss, in \aname{}, we allow different pooling methods to be composed for the $\mathbf{Q}$ and $\mathbf{K}$ tensors to better perverse their characteristics. We use the combinations of average, max, and min pooling. When appling more than one pooling methods on either $\mathbf{Q}$ or $\mathbf{K}$, the resulting pooled tensors will first be concatenated in the hidden dimension before being fed into the linear layer. 
\Figref{fig:pooling} shows the test perplexity on PG19~\citep{PG19} datasets with the best 15 pooling combinations using the Llama-3.1-8B model. Results shows that using avgpooling on $\mathbf{Q}$ and a combination of max, min, avg pooling on $\mathbf{K}$ achieves best perplexity on across different sparsity ratios. This observation might be related to the phenomenon in LLM quantization that $\mathbf{K}$ tensors tend to have more outliers. Thus, with the aid of Max and Min pooling can better extract the features in $\mathbf{K}$ after pooling. 

\begin{figure}[]
    \centering
    \includegraphics[width=0.9\linewidth]{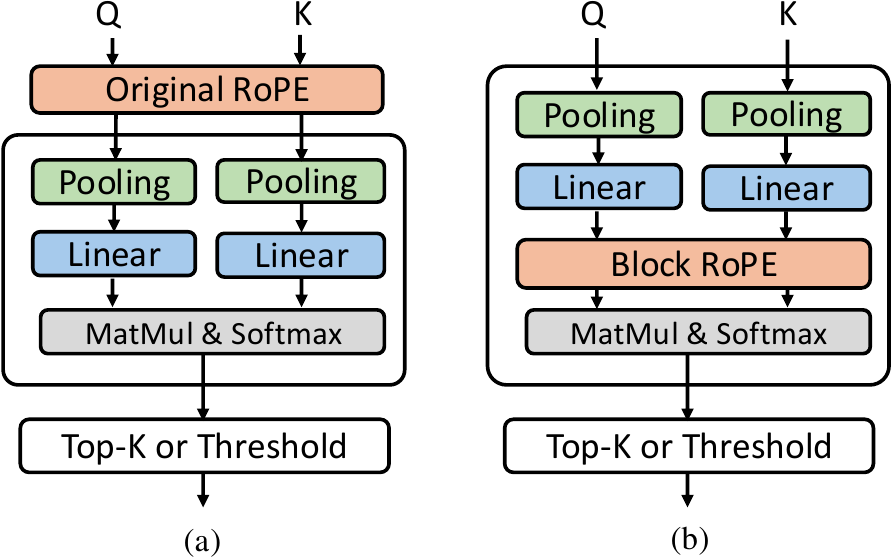}
    \caption{\textbf{Two Different RoPE Design in AttnGate.} (a) Directly taking the RoPE encoded Q and K as AttnGate input. (b) Using the Q and K before RoPE as AttnGate input and perform an block-level RoPE after linear layers. Results shows that (b) performs better and does not overfit to training data length. For simplicity, pooling composition is omitted in this diagram.}
    \label{fig:AttnGate with RoPE}
\end{figure}

\begin{figure}[h]
    \centering
    \includegraphics[width=1\linewidth]{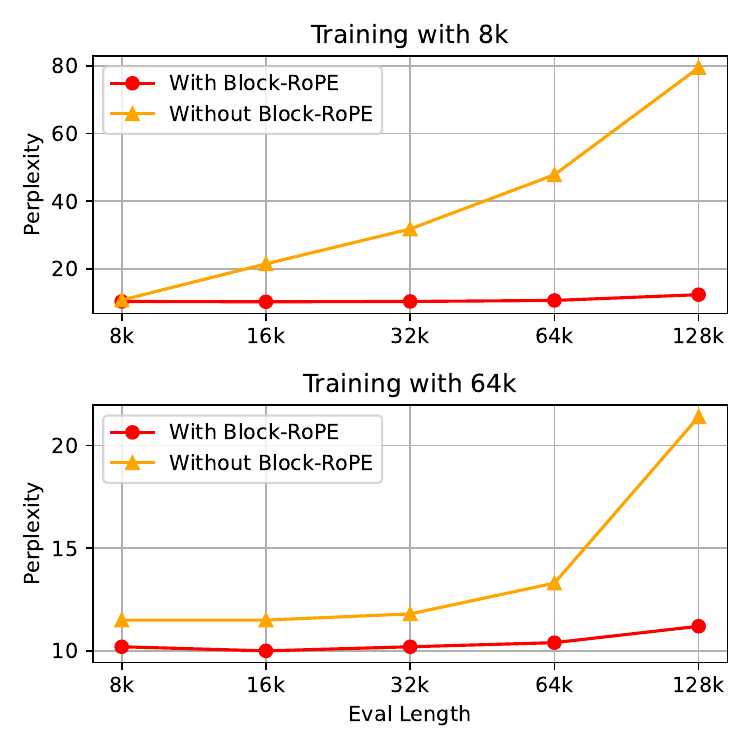}
    \caption{\textbf{Perplexity Comparison Between two RoPE design in \gate on PG19 dataset.} The block-level RoPE in \gate allows it to effectively learn the block-level positional information, resulting better test performance for different context length. The baseline design (\cref{fig:AttnGate with RoPE}a) fails to deliver reasonable results with data longer than training length.}
    \label{fig:rope results}
\end{figure}

\paragraph{Block-level RoPE in \gate.}
Modern LLMs typically employ RoPE~\citep{rope} to encode positional information. 
If the \gate relies solely on the original RoPE in the model, i.e., feeding the \gate with $\mathbf{Q}$ and $\mathbf{K}$ after RoPE, the relative positional encoding properties will be lost because of the pooling operation (shown in \Figref{fig:AttnGate with RoPE}a). 
This compromises the \gate's ability to extrapolate to longer context lengths during \gate distillation. To address this issue, we propose a different design by feeding the \gate with $\mathbf{Q}$ and $\mathbf{K}$ before RoPE encoding, and add an additional Block-level RoPE in \gate (shown in \Figref{fig:AttnGate with RoPE}b). 
To represent the block-level information, the new RoPE in \gate uses a reduced  $\theta' = \theta / B$, where $\theta$ is the original RoPE theta.

\Figref{fig:rope results} presents the test perplexity results with and without the block-level RoPE design in \gate. The results indicate that without this block-level RoPE design, \gate fails to perform adequately on evaluation data longer than 8k when trained with 8k length data. Similarly, when trained with 64k length data, it does not perform well on 128k length data. However, with the additional block-level RoPE, \gate can extrapolate to different context lengths, significantly enhancing the model performance and training efficiency.

\subsection{\gate Distillation in \aname}

While the introduced \aname{} architecture is straightforward, training presents challenges. Jointly training the gate and model from scratch, as in MoE, is costly and difficult.
Fortunately, unlike MoE, where gating network must learn expert selection from scratch, the \gate in \aname{} has a ground truth from standard attention for distillation.

\paragraph{Obtaining the Ground Truth.} We use the 2D-MaxPooled attention map from full attention as ground truth to distill \gate, as illustrated in \Figref{fig:SeerAttention}. Semantically, it means that only when all the attention score in a block is small, the 2D-MaxPooled results will be small. This is aligned with the block-sparse definition. 
However, obtaining the max-pooled attention map for training is non-trivial especially in long-context scenarios.
Modern LLMs rely on FlashAttention, which fuses operations between layers and doesn't explicitly compute the attention map. The naïve manual implementation is impractical due to quadratic memory consumption. To address this challenge, we customize an efficient kernel that directly outputs the MaxPooled attention map ground truth by modifying FlashAttention kernel but largely reuses its original computation flow. The detailed design are explained in \cref{appendix:training kernel}. 
% With this kernel, we can distill the \gate at similar cost as inferencing an LLM using FlashAttention. 

\paragraph{Loss Function.} The Kullback-Leibler divergence loss~\citep{kl} is use to distill the \gate. Since \gate uses softmax in output similar to full attention computation, the row summation of gating score will always be 1. KL-divergence loss allows the training process to focus on mimicking the attention distribution instead of absolute magnitude like Mean-square-error loss. The overall distillation process can be expressed as:
\begin{equation}
\begin{aligned}
    gt &= \mathrm{MaxPool2D}\!\Bigl(\mathrm{softmax}\!\bigl(\tfrac{QK^T}{\sqrt{d}}\bigr)\Bigr), \\
    score &= \mathrm{AttnGate}(Q, K), \\
    loss &= D_{KL}\bigl(gt \;\|\; score\bigr). \\
\end{aligned}
\end{equation}

\subsection{Inference with \aname}
After self-distillation training process, \aname can utilizes the trained \gate to generate a gating score for each block within the full attention mechanism. These scores are then used to select the final activated sparse blocks. By integrating with our Block-sparse FlashAttention kernel, \aname can achieve significant speedup for long-context prefilling while maintaining high accuracy.
\paragraph{Generating Binary Block Mask.}
\aname provides the flexibility to convert the floating-point gating scores into a final binary block mask using either the TopK or Thresholding methods. If using the TopK method, the $k$ blocks with the highest scores in each row are selected. 
\begin{equation}
b_{ij} = 
\begin{cases}
1 & \text{if } j \in \mathrm{TopK}(score_i, k).\mathrm{index}, \\
0 & \text{otherwise}.
\end{cases}
\end{equation}
Alternatively, users can apply a gating threshold to activate blocks with scores exceeding the specified threshold. 
\begin{equation}
    b = score > threshold
\end{equation}
Notably, once \gate is trained, users can adjust the TopK ratio or threshold at test time to achieve various trade-offs.

\paragraph{Block Sparse Flash-Attn Kernel.} In designing the Block Sparse Flash-Attention kernel, the block size of \gate is aligned with the tiling size used in Flash-Attention, typically 64 or 128. By doing so, we can create a customized block-sparse Flash-Attention kernel that leverages the binary block mask generated by \gate to selectively skip the I/O and computation for unactivated blocks. This approach is highly efficient on modern GPUs, as it optimizes the processing of sparse data at the block level rather than dealing with fine-grained element-wise level, leading to significant performance gains.

%% file: 04-results.tex
\section{Experiments}
In this section, we evaluate both the accuracy and efficiency of \aname. 
% For the efficiency evaluation, we present kernel-level and end-to-end latency speedup results across various sparse configurations.
In our current experiments, block-size $B$ for the \gate and sparse kernel is fixed at 64 and \gate solely applies in the prefill stage.  
%Details of the experimental setup are provided below.

\begin{figure*}[]
    \centering
    \includegraphics[width=1\linewidth]{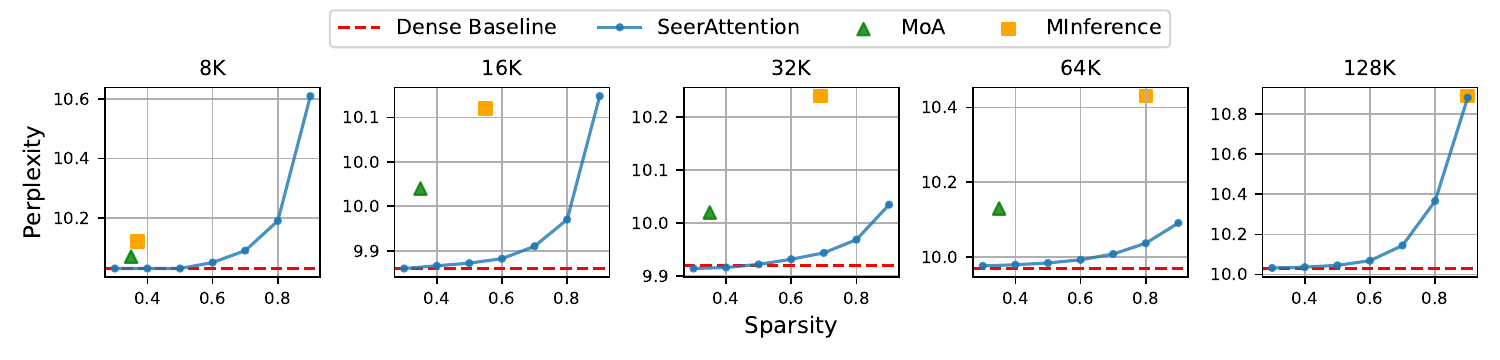}
    \caption{\textbf{Perplexity results on PG19 test split}. Each figure shows the Perplexity results of MoA, MInference, and \aname using LLama-3.1-8B-Instruct model under different evaluation context size (MoA got OOM issue on 128k evaluation). \aname allows users to adjust different sparsity ratio at test time and demonstrates better tradeoff. }
    \label{fig:ppl}
\end{figure*}

\paragraph{Models, Tasks and Baselines.} 

We apply \aname{} to the pre-trained models Llama-3.1-8B-Instruct~\citep{llama_3} in the following experiments. We evaluate the \gate performance using perplexity test on PG19~\citep{PG19}, two long context benchmarks: LongBench~\citep{longbench} and RULER~\citep{ruler}, and 4 short-context task from Open LLM Leaderboard~\citep{shottasks}: HellaSwag~\citep{hellaswag}, MMLU~\citep{mmlu}, ARC-challenge~\citep{arcc}, GSM8K~\citep{gsm8k}. 
For long-context benchmark like RULER and LongBench, we follow similar practice in \citeauthor{starattn} that only applies sparsity in context rather than question in \aname.  
We compare \aname with three state-of-the-art sparse attention methods, MoA~\citep{moa}, MInference~\citep{minference}, and DuoAttention~\citep{duo}. MoA uses an offline search scheme to apply static sparse patterns across different attention heads. In our experiment, we adopt their "KV Sparsity" in 0.5 which means "Attention Sparsity" in 0.35. 
MInference dynamically generates sparse indices using heuristic methods for each head based on pre-defined sparse patterns. We used their official configuration for Llama-3.1-8B-Instruct model, where all attention heads choose the "Vertical-Slash" sparsity pattern. DuoAttention differentiates some attention heads as streaming heads~\citep{streamingllm} while keep the rest as dense heads. In the following experiment, we adopted their official setup for Llama-3.1-8B-Instruct model with 50\% head as streaming heads. All the evaluation were run on a single A100. 
%\red{It is important to note that, in post-training experiments, \aname{} adjusts sparsity solely by using different $K$ in Top-k block selections with trained Attention Gates.}

\paragraph{Distillation Training Setup.} 
%In the Post-Training usage of \aname, we only train the additional Attention Gates without updating any original weights in the original model. 
We use the RedPajama~\citep{together2023redpajama} dataset for \gate distillation, which are chunked into 64k with BOS and EOS tokens. 
Our training employs a learning rate of 1e-3 with cosine decay. We set the global batch size to 16 and conduct training for only 500 steps, leveraging DeepSpeed~\citep{deepspeed} stage 2 optimization on A100 GPUs. As only \gate parameters are learned and updated, the distillation process can be completed within 40 A100 hours. To prevent the quadratic memory explosion that occurs when saving the intermediate attention map for ground truth generation, we customized a FlashAttention kernel. This kernel directly outputs the 2D max-pooled ground truth on top of the original attention computation. Further details about this kernel can be found in \cref{appendix:training kernel}.

% \paragraph{Long-context Extension Fine-tuning Setup.}
% We extend the context size of a Llama-3-8B model from 8K to 32K, following the setup from YaRN~\citep{yarn}, while introducing attention sparsity via \aname. The Top-k number in the \gate is fixed during the forward pass to allow the model to adapt to the sparsity.
% We use a learning rate of 1e-5 with linear decay and a global batch size of 8 on RedPajama dataset. The entire model weights are fine-tuned on 4 A100 GPUs with DeepSpeed stage 3 optimization. 

\subsection{Accuracy of Evaluation}

\paragraph{Perplexity Results.}

\Figref{fig:ppl} presents the perplexity results on the PG19 test split across various evaluation lengths. 
All documents exceeding 128k tokens from PG19 test split are selected, and the input sequences are truncated to evaluation context length before evaluation. 
We compare the performance of \aname with MoA and MInference. Notably, with a single trained \gate, users have the flexibility to adjust the TopK or threshold values during testing, allowing them to achieve different trade-offs between accuracy and efficiency. The figure illustrates the results for different TopK sparsity levels of \aname, where we uniformly apply the same TopK ratio to all attention heads. The results indicate that \aname generally offers a better trade-off compared to MoA and MInference, achieving lower perplexity at similar sparsity ratios. Note that the MoA's result for 128k is missing due to an Out-of-memory (OOM) issue on a single A100 GPU.

\paragraph{LongBench Evaluation.}

\begin{table}[]
\footnotesize
\centering
\caption{LongBench Results on Llama-3.1-8B-Instruct Model.}
\label{tab:longbench}
\begin{tabular}{cccccc}
\hline
\multicolumn{1}{c|}{} &
  0-4k &
  4-8k &
  8k+ &
  \begin{tabular}[c]{@{}c@{}}Avg.\\ Acc.\end{tabular} &
  \begin{tabular}[c]{@{}c@{}}Avg.\\ Sparsity\end{tabular} \\ \hline
\multicolumn{1}{c|}{Full Attention}  & 55.32  & 53.98  & 52.9  & 54.07 & 0.0  \\
\multicolumn{1}{c|}{MInference}      & 55.23  & 53.78  & 52.18 & 53.73 & 0.31 \\
\multicolumn{1}{c|}{MoA}             & 50.74  & 49.84  & 51.89 & 50.82 & 0.35 \\
\multicolumn{1}{c|}{DuoAttention}    & 53.77  & 52.17  & 51.27 & 52.40 & 0.5* \\
\multicolumn{1}{c|}{\textbf{SeerAttention}} &
  \textbf{55.43} &
  \textbf{54.49} &
  \textbf{52.69} &
  \textbf{54.20} &
  \textbf{0.50} \\ \hline
\multicolumn{6}{l}{* 50\% streaming heads, the real sparsity \textless{}50\%}
\end{tabular}
\end{table}
LongBench is a long-context understanding benchmark. We compare with those of MoA, MInference, and DuoAttention using the Llama-3.1-8B-Instruct model. DuoAttention uses 50\% of the heads as streaming heads and 50\% as dense heads. For streaming heads, the attention only occurs in the attention sink and recent tokens. As a result, it is not less than 50\% sparsity overall.
In this benchmark test, \aname employs a threshold of 2e-3 for all \gates. With the same threshold, different attention gates can exhibit varying sparsity ratios, and longer context data tends to be sparser. This approach allows for a more adaptive allocation of sparsity.
As demonstrated in \tabref{tab:longbench}, \aname consistently outperforms other methods across various test lengths. Notably, in the 0-4k and 4-8k tests, our score surpasses even the dense baseline.  This may be attributed to \gate filtering out noisy attention in certain cases. Furthermore, \aname achieves the highest average score and the highest average sparsity across all tests.

\begin{table*}[]
\centering
\caption{RULER Benchmark Results on Llama-3.1-8B-Instruct Model.}
\label{tab:ruler}
\begin{tabular}{ccccccccc}
\hline
Methods &
  4k &
  8k &
  16k &
  32k &
  64k &
  128k &
  \begin{tabular}[c]{@{}c@{}}Average\\ Accuracy\end{tabular} &
  \begin{tabular}[c]{@{}c@{}}Average\\ Speedup\end{tabular} \\ \hline
Full Attention & 95.53          & 92.37 & 92.01 & 87.63 & 84.39 & 76.26          & 88.01 & 1.00 \\
MInference     & 95.53          & 92.64 & 91.37 & 85.71 & 83.24 & 67.02          & 85.92 & 0.83 \\
DuoAttention   & \textbf{95.64} & 92.08 & 90.71 & 84.75 & 83.24 & \textbf{75.32} & 86.96 & 1.09 \\
\textbf{SeerAttention} &
  95.53 &
  \textbf{92.71} &
  \textbf{92.02} &
  \textbf{88.49} &
  \textbf{83.48} &
  73.37 &
  \textbf{87.60} &
  \textbf{1.41} \\ \hline
\end{tabular}
\end{table*}

\paragraph{RULER Evaluation.}
RULER is a long-context LLM evaluation benchmark consisting of 13 challenging sub-tasks. It generates tests with data sizes ranging from 4k to 128k. In this experiment, \aname employs a threshold of 5e-4, which allows it to automatically adapt sparsity from approximately 10\% for 4k data to around 85\% for 128k data.
Due to out-of-memory (OOM) issues in some tests, MoA was excluded from this benchmark. Table \ref{tab:ruler} provides detailed accuracy results across different evaluation lengths. It is evident that \aname achieves the best accuracy in most tests (8k-64k). For the 128k test, DuoAttention has less than 50\% sparsity, whereas \aname maintains an sparsity higher than 80\%, which accounts for the slightly lower performance.
\aname also attains the highest average accuracy compared to other models (only 0.41\% lower than the dense baseline) while delivering the highest average end-to-end speedup ($1.41\times$) in prefilling time.

\begin{table}[]
\footnotesize
\centering
\caption{Short Context Tests on Llama-3.1-8B-Instruct Model}
\label{tab:short task}
\begin{tabular}{ccccc}
\hline
                   & MMLU & HellaS.   & ARC-c & GSM-8K \\ \hline
Full Attention & 68.1 & 80.1 & 60.7  & 75.7   \\ \hline
SeerAttention  & 67.9 & 79.8 & 60.2  & 75.6   \\
Avg Sparsity       & 3.4  & 50.4 & 26    & 52.1   \\
Avg Seqlens        & 118  & 840  & 395   & 872    \\ \hline
\end{tabular}
\end{table}
\paragraph{Short Context Test.}
For short context input, attention contributes a smaller proportion in the total runtime. Consequently, sparse attention does not significantly enhance latency performance. Nevertheless, we evaluate \aname accuracy performance under a very high threshold 3e-2 to achieve high sparsity. The results, as shown in \cref{tab:short task}, indicate that \aname exhibits negligible accuracy loss. For instance, with an average sequence length of 872 in the GSM-8K task, \aname achieves only 0.1\% degradation in accuracy with 52\% averaged sparsity.

\begin{figure*}[t]
    \centering
    \includegraphics[width=1\linewidth]{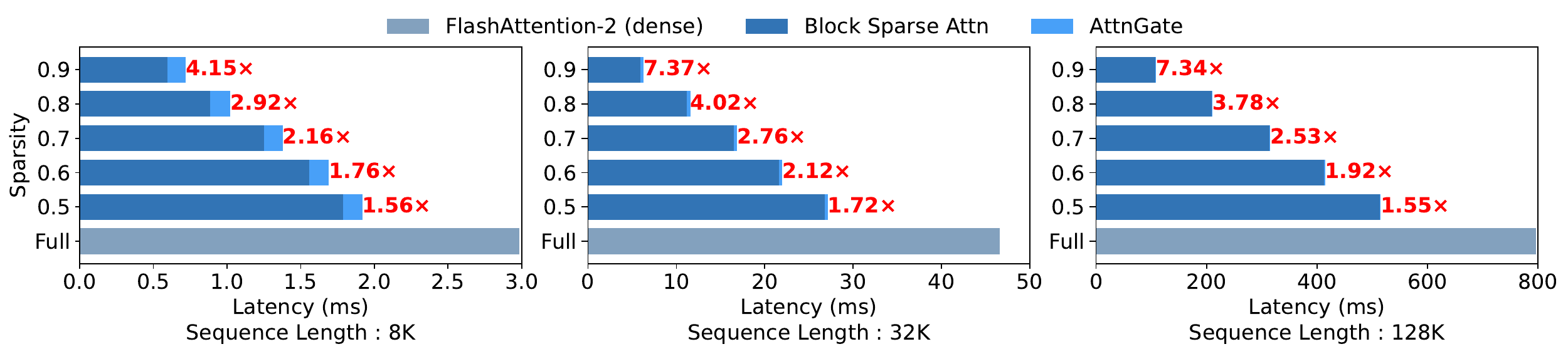}
    \captionsetup{skip=0pt}
    \caption{\textbf{\aname Speedup over FlashAttention-2 at the Kernel Level.} The latency overhead from \gates is minimal. Our block-sparse attention kernel achieves highly linear speedup over dense configurations, delivering a 7.3× speedup with 90\% sparsity on sequences of 128k. The \gate overhead amost diminishes in 128k context length.}
    \label{fig:kernel breakdown}
\end{figure*}

\subsection{Efficiency Evaluation}
% We evaluate the efficiency of \aname using our implementation of Triton~\citep{triton} kernels. We evaluate the kernel-level as well as end-to-end speedup using a Llama-3.1-8B-Instruct on a single A100 GPU. Results are compared to FlashAttention-2 (dense baseline), MoA and MInference. 

We evaluate the efficiency of \aname using our implementation of CUDA kernels. We evaluate the kernel-level as well as end-to-end speedup using a Llama-3.1-8B-Instruct on a single A100 GPU. Results are compared to FlashAttention-2 (dense baseline), MoA, MInference and DuoAttention. 

\subsubsection{Kernel evaluation}

\paragraph{Negligible Overhead incurred by \gate.}
% In the inference pre-filling stage, \aname integrates an Attention Gate, Top-k filtering, and Block-Sparse Attention mechanisms. Section 3.2 provides a comprehensive overview of the \aname model architecture, where we employ Triton to facilitate the implementation of a fused attention gate kernel and leverage the naive PyTorch Top-k operation to extract row-wise Top-k block indices. 
\cref{fig:kernel breakdown} shows the kernel-level latency breakdown of \aname. It demonstrates that the overhead introduced by the \gate during inference is minimal. For instance, at a context length of 32K and a sparsity of 0.5, the \gate contributes only 1\% to the total latency of an attention layer. In the cases of 128K sequence length, the relative overhead almost diminishes.

\paragraph{Block-sparse FlashAttention Kernel Speedup.}
%We evaluate our own implementation of Block-Sparse FlashAttention kernel that only processes sparse blocks identified by \gate. 
\Figref{fig:kernel breakdown} also shows that our kernel exibits linear speedup over various sparsity levels. At a sequence length of 128K with 90\% sparsity, \aname achieves a speedup of $7.3\times$ compared with FlashAttention-2 (full attention) on a single A100 GPU. This demonstrates the effectiveness of the block-level sparsity employed by \aname, which is highly efficient on GPUs and translates into high speedup.

\begin{figure}[]
    \centering
    \includegraphics[width=\linewidth]{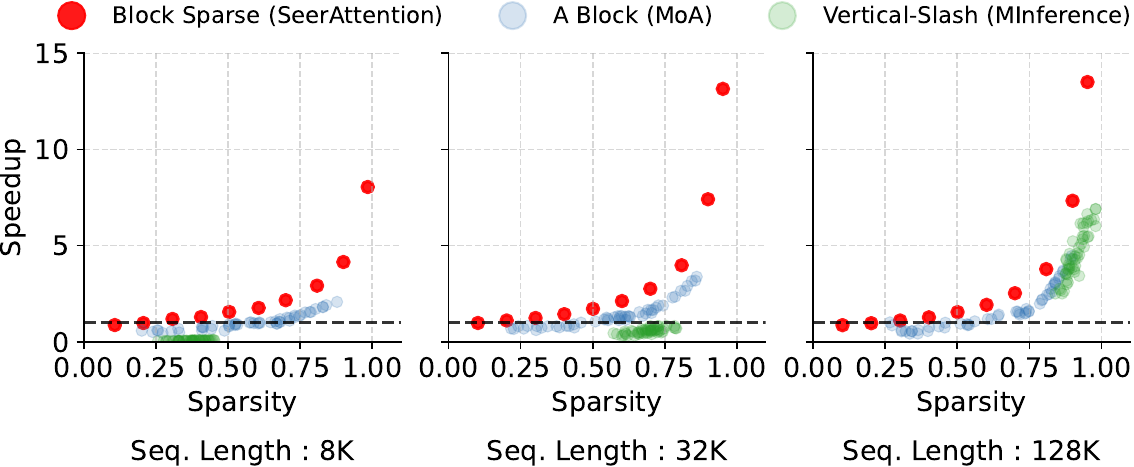}
    \caption{Kernel-level Speedup Comparison Between Different Works. \aname translates sparsity to speedup more effectively. }
    \label{fig:layer}
\end{figure}

\begin{figure*}[t]
    \centering
    \begin{subfigure}[b]{0.19\linewidth}
        \centering
        \includegraphics[width=\linewidth]{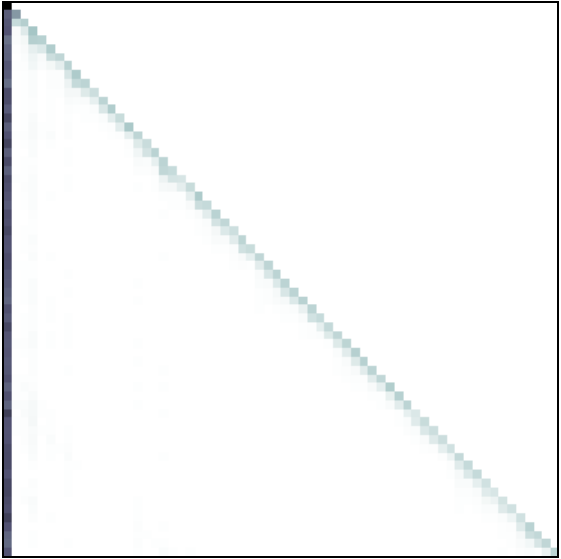}
        \caption{}
        \label{fig:mask_0}
    \end{subfigure}
    \hfill
    \begin{subfigure}[b]{0.19\linewidth}
        \centering
        \includegraphics[width=\linewidth]{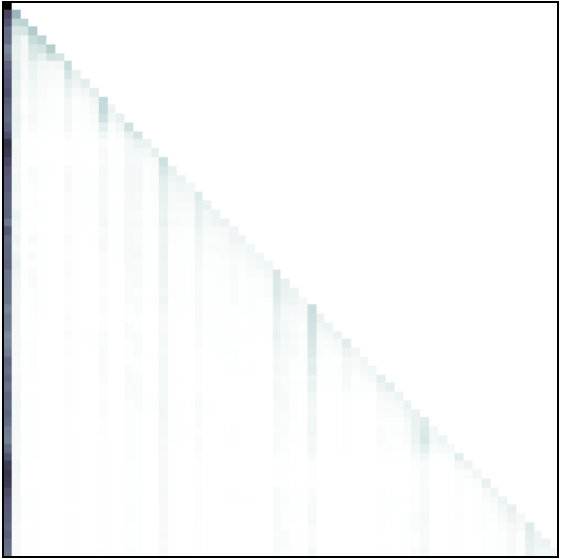}
        \caption{}
        \label{fig:mask_1}
    \end{subfigure}
    \hfill
    \begin{subfigure}[b]{0.19\linewidth}
        \centering
        \includegraphics[width=\linewidth]{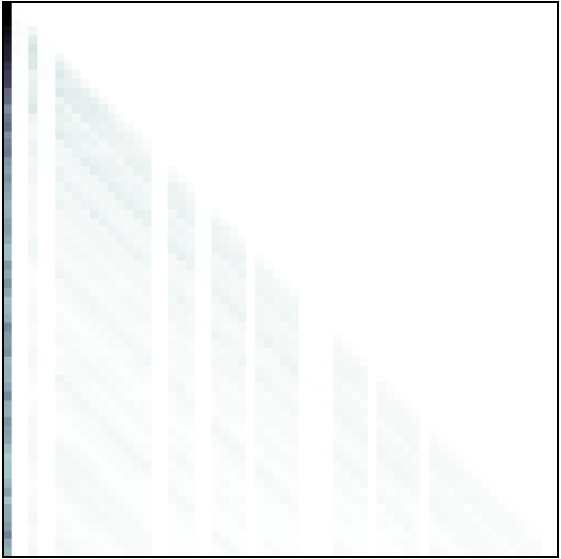}
        \caption{}
        \label{fig:mask_2}
    \end{subfigure}
    \hfill
    \begin{subfigure}[b]{0.19\linewidth}
        \centering
        \includegraphics[width=\linewidth]{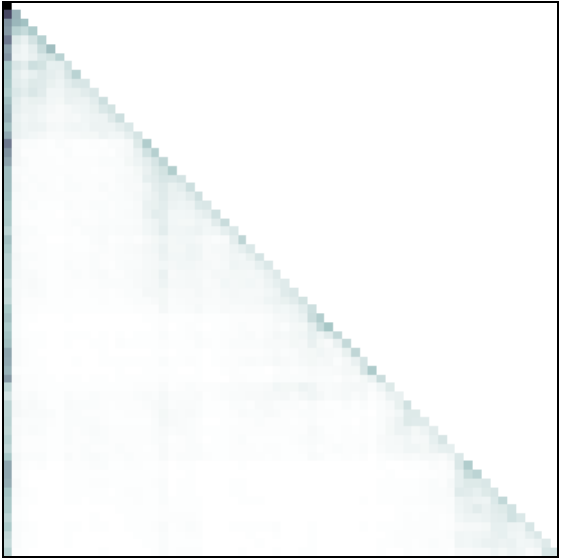}
        \caption{}
        \label{fig:mask_3}
    \end{subfigure}
    \hfill
    \begin{subfigure}[b]{0.19\linewidth}
        \centering
        \includegraphics[width=\linewidth]{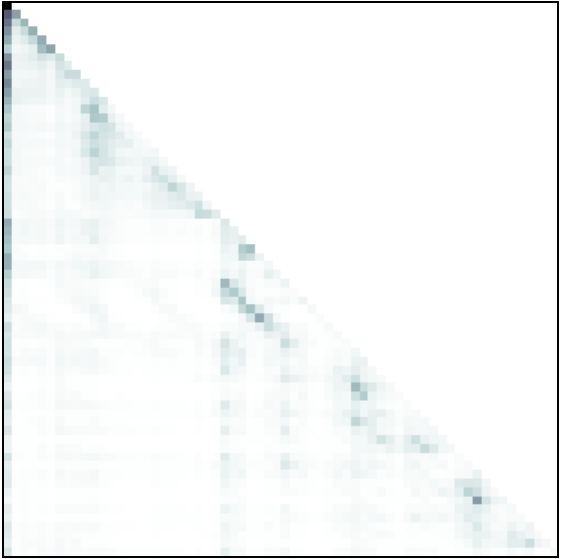}
        \caption{}
        \label{fig:mask_4}
    \end{subfigure}
    \caption{Visualization of the \gate's outputs.}
    \label{fig:vis mask}
\end{figure*}

\paragraph{Kernel-level Comparison with Related Works.}
We compare the kernel-level speedup of \aname with MoA and MInference. MInference uses offline calibration to identify a pre-defined sparse pattern for each layer. For Llama-3.1-8B-Instruct model, MInference consistently uses "Vertical-slash" pattern across all layers. During runtime, MInference will dynamically generate non-zero indices based on their approximation algorithm. On the other hand, MoA uses "A-shape" blocks as their sparse pattern and calibrate the shape parameters offline under given sparsity constraint.  DuoAttention is omitted in kernel-level comparison as it's a combination between streaming and dense head, whose performance is a mixture results of block sparse attention and dense FlashAttention. 

\Figref{fig:layer} shows the sparsity v.s. speedup plots of different methods on 8k, 32k, 128k sequences length, where the speedup baseline is FlashAttention-2. 
The kernel-level sparsity statistics were collected from PG19 datasets. For MoA, we generated the sparse configurations under their 0.5 overall "KV-sparsity" constraints, which corresponds to an average of 0.35 sparsity in attention. The results demonstrates that the block-sparse attention kernel used in \aname outperforms both MoA and MInference in most cases.

\begin{figure}[t]
    \centering
    \includegraphics[width=\linewidth]{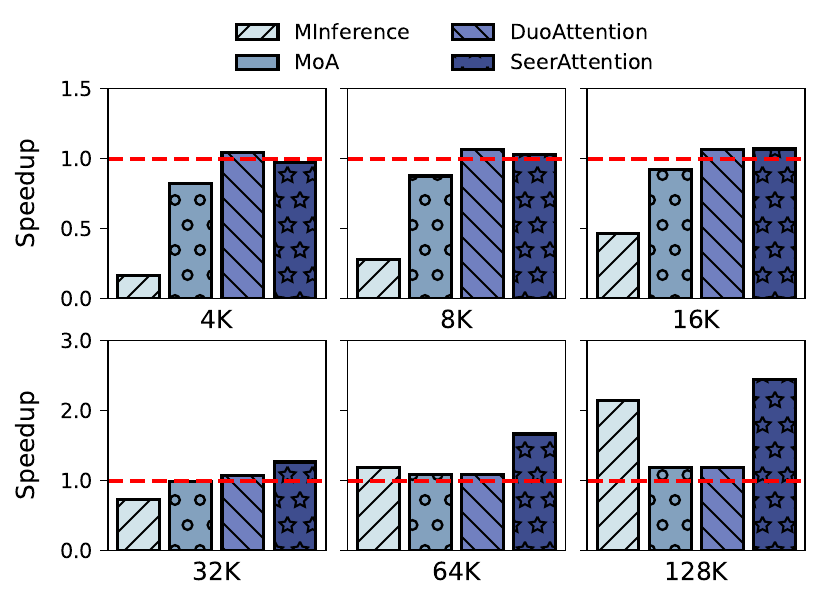}
    \caption{\textbf{Comparing Prefilling Time Speedup on RULER Test Setting.} \aname outperforms related works in most long-context data scenarios ($\geq$ 16k). For longer context data, the attention mechanism constitutes a larger proportion of the total runtime, allowing sparse methods to achieve better speedup. Overall, \aname achieves the highest average speedup ($1.41\times$) while maintaining the best average accuracy under this RULER benchmark setting. }
    \label{fig:end2end speedup}
\end{figure}

% \begin{figure}[ht]
%     \centering
%     \begin{subfigure}[b]{0.33\textwidth}
%         \centering
%         \includegraphics[width=\textwidth]{figures/kernel_8K.pdf}
%         \caption{Sequence length = 8k}
%         \label{fig:kernel_8k}
%     \end{subfigure}
%     \hfill
%     \begin{subfigure}[b]{0.326\textwidth}
%         \centering
%         \includegraphics[width=\textwidth]{figures/kernel_32K.pdf}
%         \caption{Sequence length = 32k}
%         \label{fig:kernel_32k}
%     \end{subfigure}
%     \hfill
%     \begin{subfigure}[b]{0.318\textwidth}
%         \centering
%         \includegraphics[width=\textwidth]{figures/kernel_128K.pdf}
%         \caption{Sequence length = 128k}
%         \label{fig:kernel_128k}
%     \end{subfigure}
%     \caption{\aname block sparse FlashAttention inference kernel speedup.}
%     \label{fig:compared}
% \end{figure}

\paragraph{End-to-end Speedup Comparison.}

% To show the practical speedup of \aname, we deploy the framework into real-world single-batch scenarios based on Llama-3.1-8B. We measure the average latency of time to first token (TTFT) in the prefilling stage under different sequence lengths and sparsity ratio. We compare \aname, MoA, MInference with a full attention baseline which is implemented by FlashAttention-v2. Similarly, we record the average sparsity ratio of MInference in different context length ranges. Results in Table ~\ref{tab:ttft} shows that SeerAttention achieves lower latency than MoA and MInference under similar sparsity ratios. MoA even cannot handle over 64K context length since they store "indices look-up table  mask" for all attention layers and cause out of memory. At context length of 64k with 0.8 sparsity, \aname boosts inference speed by 1.31x compared with FlashAttention-v2 implementation.

To assess the end-to-end speedup of our method, we measured the average prefilling time, or time-to-first-token (TTFT), using the Llama-3.1-8B-Instruct model on the RULER test discussed above. Since attention takes up more runtime with longer contexts, all methods generally achieve better speedup with longer context lengths.
It should be noted that \aname uses an identical threshold across all tests in RULER,  automatically adjusting to higher sparsity for longer contexts (ranging from approximately 10\% sparsity for 4k to around 85\% sparsity for 128k). This approach results in an end-to-end prefilling speedup of up to $2.43\times$ on 128k length. On the other hand, MInference experiences a slowdown with data sizes less than 64k due to significant overhead in searching for sparse indices during runtime.
It is feasible for \aname to adjust to higher sparsity to achieve greater speedup in shorter contexts as a tradeoff. Nevertheless, \aname delivered the highest average accuracy (87.6) and the greatest average speedup ($1.41\times$) in this RULER benchmark setting.

\subsection{Visualization of Learned Attention Maps.}
The \gate module automatically learns diverse sparse patterns without any prior knowledge or heuristics. \Figref{fig:vis mask} shows several example outputs from \gate, including (a) "A-shape," or streaming head  (b) "Vertical," (c) "Slash" with empty vertical spaces, (d) block sparsity along the diagonal, and (e) random patterns. These patterns not only encompass but also extend beyond those observed in previous works such as MoA and MInference, showcasing the versatillty of our learning based methods.

%% file: 05-conclusion.tex
\section{Conclusion and Future Work}
This paper presents \aname{}, a new attention mechanism that learns and leverages the intrinsic sparsity in attention to boost long-context LLMs.
\aname{} learns the attention sparsity from the LLM itself with a lightweight self-distillation approach.
Our experiments demonstrate that \aname{} outperforms previous approaches in terms of long context model accuracy and pre-filling latency.
For future work, there are several promising directions to explore for improving and expanding the capabilities of \aname.
One key area is enhancing the training methodologies for \aname, such as applying \aname in long-context continued pre-training with more training tokens to achieve higher sparsity without sacrificing accuracy (preliminary experiments in \cref{appendix:preliminary finetune}).
%shijie: add finetuning in appendix?
Another important avenue is applying \aname{} in the decoding stage, especially for long-CoT.

%% file: 00-appendix.tex
\section{Appendix}
\subsection{Training \aname with Customized GPU Kernel}
\label{appendix:training kernel}

% In Section 3.2, we discussed the method for obtaining the ground truth attention map used to distill \gate. Specifically, we leverage the 2D-MaxPooled attention map from full attention as the ground truth, aligning with the block-sparse attention definition. However, directly computing this attention map is challenging due to the quadratic memory complexity and the fused operation nature of FlashAttention. To overcome this, we developed a customized kernel that efficiently extracts the max-pooled attention map by modifying the FlashAttention kernel while largely preserving its computation flow.

In this appendix, we provide a detailed design and implementation of our efficient kernel, highlighting key modifications to FlashAttention and optimizations for long-context scenarios. We then evaluate the peak memory usage and additional latency overhead of our training kernel during the AttnGate training stage, showing that it incurs only minimal overhead in both memory and latency compared to training with FlashAttention-2.

\paragraph{FlashAttenion with  2D-MaxPooling: A Customized training kernel.}

\begin{figure}[h]
    \centering
    \includegraphics[width=0.74\linewidth, page=3]{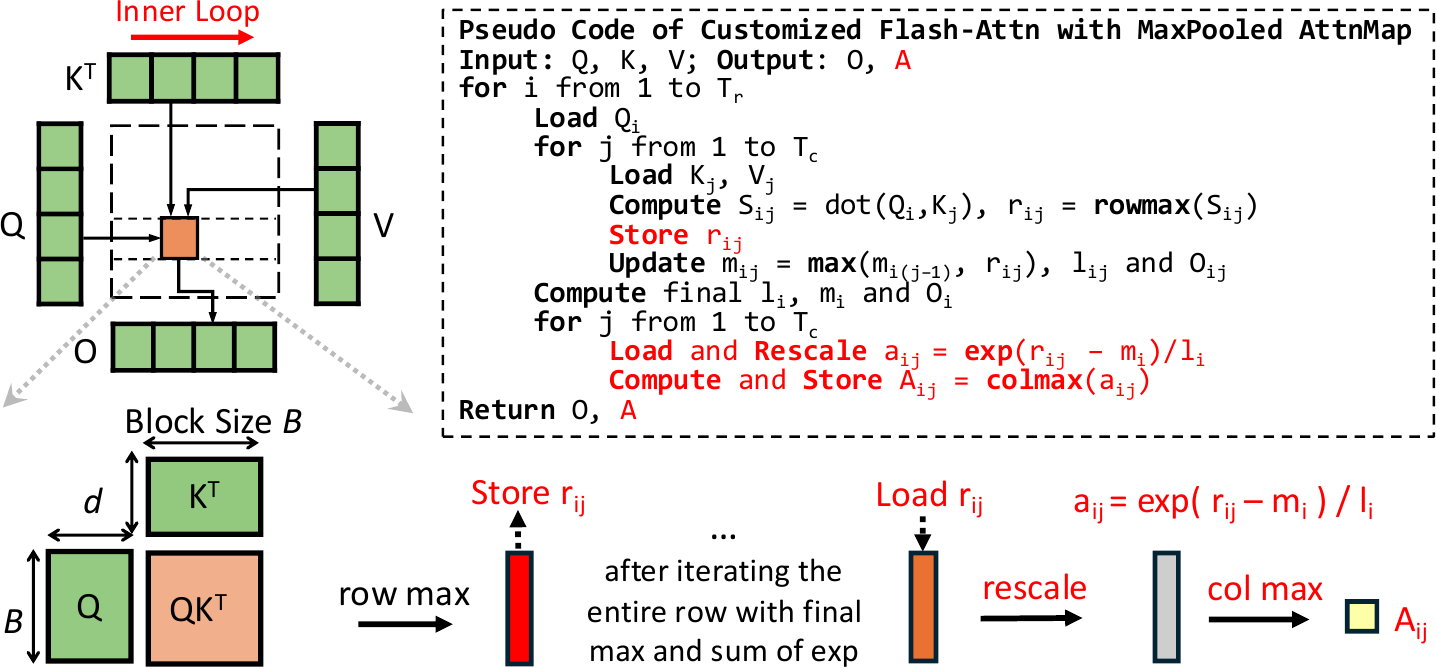}
    \caption{Efficient FlashAttention kernel with pooling of attention map. }
    \label{fig:attn_pooling}
\end{figure}

In Section 3.2, we discussed the method for obtaining the ground truth attention map used to distill \gate. Specifically, we leverage the 2D-MaxPooled attention map from full attention as the ground truth, aligning with the block-sparse attention definition. However, directly computing this attention map is challenging due to the quadratic memory complexity and the fused operation nature of FlashAttention. To overcome this, we developed a customized kernel based on Triton \cite{triton} that efficiently extracts the  2D-MaxPooled attention map by modifying the FlashAttention kernel while largely preserving its computation flow. \Figref{fig:attn_pooling} shows the pseudo code and diagram of this customized kernel.
%For instance, attention maps (16-bit precision) for a model with 32 attention layers, each with 32 heads, and 8k sequence length requires $(8k)^2 \times 32 \times 32 \times 2 = 128$ GB of GPU memory—far exceeding the capacity of an A100 GPU.

\begin{figure}[h]
    \centering
    \begin{subfigure}[b]{0.4\textwidth}
        \centering
        \includegraphics[width=\textwidth]{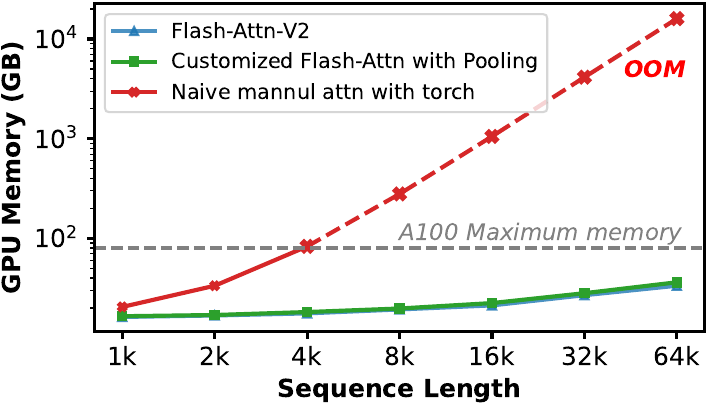}
        \caption{Memory}
        \label{fig:pooling_sub1}
    \end{subfigure}
    \hspace{0.05\textwidth}
    \begin{subfigure}[b]{0.4\textwidth}
        \centering
        \includegraphics[width=\textwidth]{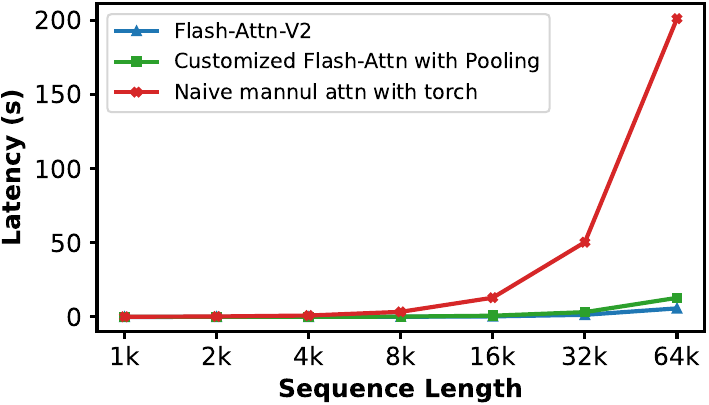}
        \caption{Latency}
        \label{fig:pooling_sub1}
    \end{subfigure}
    \caption{Memory and latency of customized FlashAttention with max-pooling training kernel.}
    \label{fig:pooling_mem_latency}
\end{figure}

Normally, the softmax function ensures numerical stability by subtracting the maximum value before applying the exponential operation. FlashAttention computes the local row max of each block, and gradually updates the global maximum through iteration:
\begin{equation}
\begin{split}
    S_{ij} &= Q_i K^T_j; \\
    r_{ij} &= \mathrm{rowmax}(S_{ij}); \\
    m_{ij} &= \max(m_{i(j-1)}, r_{ij}).
\end{split}
\end{equation}

where $r_{ij}$ is typically treated as a temporary result. However, we store it in HBM and rescale it later with the final global max $m_i$ and sum of exp $l_i$ after the iteration:
\begin{equation}
    a_{ij} = \mathrm{exp}(r_{ij} - m_i) / l_i
\end{equation}
This $a_{ij}$ represents the correct row max of the original attention block. With that,  2D-MaxPooling is achieved by applying a column max over $a_{ij}$. This introduces only minor overhead (storing and rescaling $r_{ij}$) but significantly improves the efficiency of obtaining the ground truth. The overhead of memory and latnecy analysis is in Figure~\ref{fig:pooling_mem_latency}.

\paragraph{Performance of the Training Kernel.}
We evaluate our customized FlashAttention kernel with  2D-MaxPooled attention map for scalable training of \aname by comparing against with PyTorch na\"ive manual attention implementation and FlashAttention-2. As shown in Figure \ref{fig:pooling_sub1}, the PyTorch kernel runs out of memory (OOM) when the sequence length exceeds 4k, while our customized kernel costs similar peak memory usage compared to FlashAttention-2. Regarding latency, since PyTorch encounters OOM for sequences longer than 8K, the attention operations per head into a loop to assess kernel-level latency. Figure \ref{fig:pooling_sub1} shows that the latency overhead introduced by the additional pooling operation is minimal compared to FlashAttention-2, while the PyTorch implementation suffers from a significant slowdown.

\subsection{Preliminary Experiments of Fine-tuning with \aname}
\label{appendix:preliminary finetune}

\begin{figure}[h]
    \centering

        \includegraphics[width=0.35\textwidth]{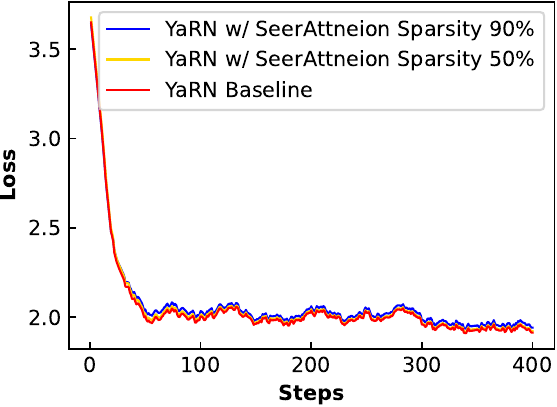}
        \caption{Fine-tuning Loss.}
        \label{fig:sub1}

    \caption{
    By incorporating \aname with YaRN~\citep{yarn} to extend a Llama-3-8B model from 8k to 32k context length, the loss curves for 50\% to 90\% sparsity are nearly identical to the dense YaRN baseline.}
    \label{fig:teaser}
\end{figure}

\begin{table}[h]
\centering
\footnotesize
\caption{Perplexity of YaRN baseline, \aname after YaRN and YaRN fine-tuning with \aname.}
\label{tab:continued pretraining}
\begin{tabular}{c|c|ccccc|ccccc}
\hline
 & YaRN  & \multicolumn{5}{c|}{Post-training \aname after YaRN} & \multicolumn{5}{c}{YaRN with \aname} \\
Sparsity   & 0.0  & 0.5  & 0.6  & 0.7  & 0.8  & 0.9   & 0.5  & 0.6  & 0.7  & 0.8  & 0.9  \\ \hline
PG19       & 8.79 & 9.16 & 9.30 & 9.48 & 9.73 & 10.18 & 8.81 & 8.82 & 8.85 & 8.93 & 9.16 \\
Proof-pile & 2.46 & 2.53 & 2.57 & 2.61 & 2.68 & 2.85  & 2.47 & 2.47 & 2.48 & 2.51 & 2.60 \\ \hline
\end{tabular}
\end{table}

In this preliminary experiment, we demonstrate that \aname can be seamlessly integrated in Long-context extension fine-tuning stages. 
We follow YaRN~\citep{yarn} to extend the context size of a Llama-3-8B model from 8k to 32k. The loss function is the summation of original cross-entropy loss and \gate loss.  To ensure stable training, the
\gates are first initialized using the post-training self-distillation before fine-tuning the entire model.
We integrate \aname into YaRN and compare the performance against the YaRN dense baseline and the post-training time self-distillation of \aname applied after YaRN.
\Figref{fig:sub1} presents the loss curves of the YaRN dense baseline and \aname at 50\% and 90\% sparsity. The curve at 50\% sparsity nearly overlaps with the baseline, while the curve at 90\% sparsity shows slightly higher loss.
\tabref{tab:continued pretraining} displays the test perplexity on the PG19 and ProofPile datasets evaluated at a 32k context length. The YaRN dense baseline achieves perplexity scores of 8.79 and 2.46, respectively. Post-training \aname results in increased perplexity. 
%for instance, at 50\% sparsity, perplexity rises from 8.79 to 9.16 and from 2.46 to 2.53.
When applying \aname during the YaRN extension fine-tuning, it maintains near-lossless performance at 50\% sparsity (with scores of 8.81 and 2.47), and even at 90\% sparsity, the loss remains minimal.
%demonstrating the capability and potential of \aname to be applied across a broader range of the LLM lifecycle. 
%\Figref{fig:Pooling_exp} shows the loss curves of the baseline model and the model with 50\% and 90\% sparsity. We can see that the curve with 50\% sparsity nearly overlaps with the baseline curve, while the curve with 90\% sparsity shows slightly higher loss than the baseline throughout the training.